\documentclass[12pt,journal,onecolumn]{IEEEtran}

\usepackage{amsmath,amssymb,amsfonts,amsthm}
\usepackage{cite}
\usepackage{url}
\usepackage{graphicx}

\begin{document}

\title{New Foggy Object Detecting Model}

\author{
Rahul~Banavathu,~Modem~Veda~Sree,~Bollina~Kavya~Sri, and~Suddhasil~De~(\IEEEmembership{Senior~Member,~IEEE})\\%
Department of Computer Science and Engineering,\\%
National Institute of Technology Patna,~India\\%
Email: suddhasil.de@acm.org%
}

\maketitle

\thispagestyle{empty}
\pagestyle{empty}

\begin{abstract}
Object detection in reduced visibility has become a prominent research area. The existing techniques are not accurate enough in recognizing objects under such circumstances. This paper introduces a new foggy object detection method through a two-staged architecture of region identification from input images and detecting objects in such regions. The paper confirms notable improvements of the proposed method's accuracy and detection time over existing techniques.
\end{abstract}


\section{Introduction}\label{sec:Intro}

Self-driving vehicles are going revolutionary in everyday life~\cite{enwiki:1197134014}. But it becomes a significant challenge when the weather conditions are not ideal, e.g., foggy conditions~\cite{dri24}. Such circumstances require technological advancement in order to function accordingly, otherwise chances of disaster are impending~\cite{pandey2017communication}. Despite the availability of several robust solutions, it becomes quite an issue in terms of accuracy while being constrained over speed.

Literature has advocated Region-based Convolutional Neural Network (R-CNN)~\cite{enwiki:1192366922} for detecting objects. R-CNN has initial objective to receive any input image and produces as output a collection of bounding boxes, everyone of which would include an object and corresponding category to indicate nature of that object. With time, R-CNN has been expanded to handle additional tasks of computer vision. The first expansion of R-CNN is to extract regions of interest (ROI)~\cite{enwiki:1113957817} from input through Region Proposal Network (RPN), and subsequently, a neural network~\cite{enwiki:1198631061} is fed with each ROI to generate its output features. Then, set of support-vector machine (SVM) models~\cite{enwiki:1190739318} is employed for each ROI's output features to identify the object and its category (if any) that is present within that ROI. An alternate direction of expanding R-CNN is fast R-CNN (FRCNN), in which after ROIs are extracted by RPN, the neural network is executed only on the entire set of ROIs, which results into computed features of all ROIs. ROI-pooling~\cite{enwiki:RoI_pooling} takes output feature set tensor of the network, slices off each ROI, restructures every ROI, and then classifies each of them. Yet another direction is faster R-CNN (FerRCNN), ROI synthesis is integrated within neural network, which also computes features of ROIs.

Literature has also advocated a revolutionary concept of domain adaptation (DA)~\cite{enwiki:1179510711} to be used in conjunction with R-CNNs. In R-CNN plus DA, labeled images from source domain is being used for training the RCNN's backbone network model. Thus, the network is capable to adapt to specific features and patterns as per ROIs. On availability of raw images from target domain, which being related to source domain, generated ROIs from such images are detected by the trained network and are further predicted into corresponding categories.

However, combining the two techniques of DA with R-CNN or its variants for object detection under foggy conditions have not produced accurate bounding boxes, or proper classifying objects in ROIs. Foggy images produce reduced visibility caused by complex atmospheric conditions caused (like, haze, fog, and smoke), as well as variations in the density of fog.

In this paper, a new foggy object detection method is proposed, which incorporates a two-stage architecture. The two stages include FerRCNN technique and DA technique. The FerRCNN technique ensures the proposed method's focus towards the detection requirement. The domain specific ROIs are crucially treated by the architecture, which discards the unwanted regions and saves time without compromising accuracy. The proposed method surpasses existing R-CNN models in terms of accuracy and detection time. Further, for real-time applications, the proposed method also reduces detection time of foggy images.

\section{Proposed method}\label{sec:Proposed}

The proposed method is an object detection technique under reduced visibility, that can detect multiple and different kinds of objects under foggy weather conditions that are present in the input image. The proposed method features two-stage domain-specific detection, which fine tunes the detection and predicts class probabilities for each kind of object. The two stages include FerRCNN technique and DA technique. The FerRCNN technique ensures the proposed method's focus towards the detection requirement. The domain specific ROIs are crucially treated by the proposed method, which discards the unwanted regions and saves time without compromising accuracy. The proposed method is trained with specific foggy datasets, that consists of images with annotations that are captured in various adverse foggy conditions, which enables the model to determine specific features that are posed by foggy environments. In the evaluation of the proposed method as two-stage object detection method, focus is kept on accuracy.
\begin{figure}[b]
\centering
\includegraphics[width=0.8\linewidth]{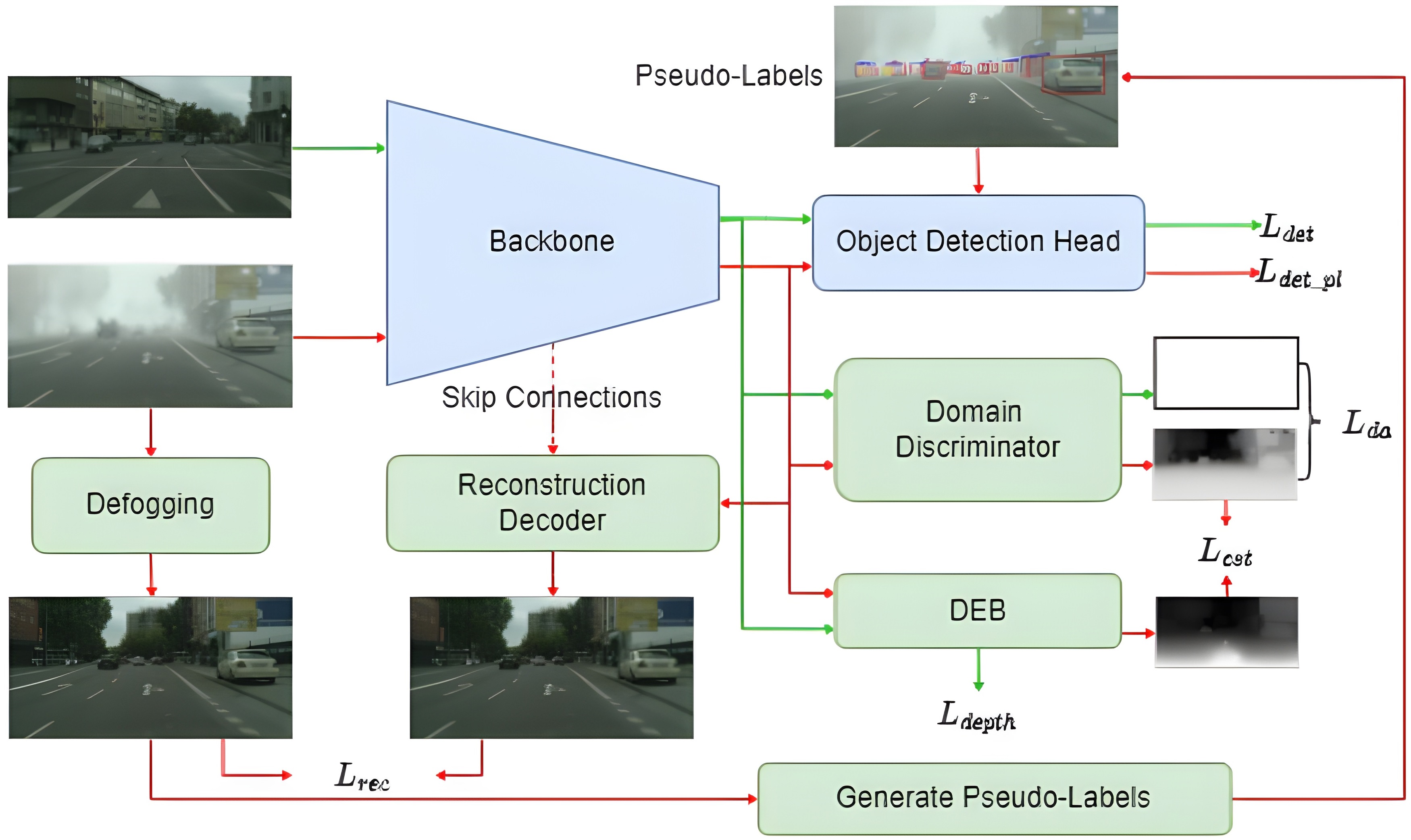}
\caption{Architecture of the proposed method.}\label{fig:prop_archi}
\end{figure}

The architecture of the proposed method is clearly portrayed by Figure~\ref{fig:prop_archi}. Here, the architecture of the proposed method builds on convolutional neural network (CNN){\footnote{Use of CNN~\cite{enwiki:1197047138} is widespread~\cite{dhanraj2019efficient}, and so CNN becomes the core of the proposed method.}}, and it is comprised of two major components, viz. the backbone network, and the head network. The backbone network extracts feature maps from the input images. The feature extraction approach starts with all ROIs being earmarked in every image by the network itself, followed by computing all features of all ROIs. ROI-pooling has the role of taking output feature set tensor of the network, slices off each ROI, restructures every ROI, and then classifies each of them. The loss function of the network is the cumulative effect of the loss term while regressing in defining bounding boxes of ROIs and the loss term while predicting present of objects in ROIs.

Using the feature maps, the object detection head locates object instances and classifies each one of them. Techniques such as domain discriminator and depth estimation block (DEB) provide the extraction of fog-invariant features through the backbone as well as preservation of depth distributions. The proposed method also incorporates reconstruction decoder and pseudo-label generator. The reconstruction decoder reduces the amount of counterfeit object characteristics produced due to DA. Target domain information is included in the 2-staged architecture by virtue of the pseudo-label generator, which also applies consistency regularization across images with and without fog. A regulating term{\footnote{Significance of regulating loss is always prioritized in the literature~\cite{kumar2023robotic}.}} for reducing loss is added in the loss function.

\section{Experiments \& Results}\label{sec:exp_res}

The experimentations of the proposed method described in Section~\ref{sec:Proposed} are performed for training, validation and testing on the benchmark dataset of images of roads and vehicles therein under foggy weather conditions. PyTorch, in conjunction with TensorCUDA, are used to perform the experiments.

\subsection{Setup}
\begin{table}[tb]
\caption{Parameters of the proposed method.}\label{tab:expsettings}
\begin{center}
\begin{tabular}{@{}l@{~} @{~}l@{}}
\hline
$\qquad$Params & $\quad$Vals$\quad$\\%
\hline
      $\quad$Train, Valid, Test data & $\quad$80\%, 10\%, 10\% of dataset\\
      $\quad$Epochs & $\quad$10\%\\
      $\quad$Batch size & $\quad$10\\
      $\quad$Number of classes & $\quad$6\\
\hline
\end{tabular}
\end{center}
\end{table}
\begin{figure}[b]
\centering
\includegraphics[width=0.6\linewidth]{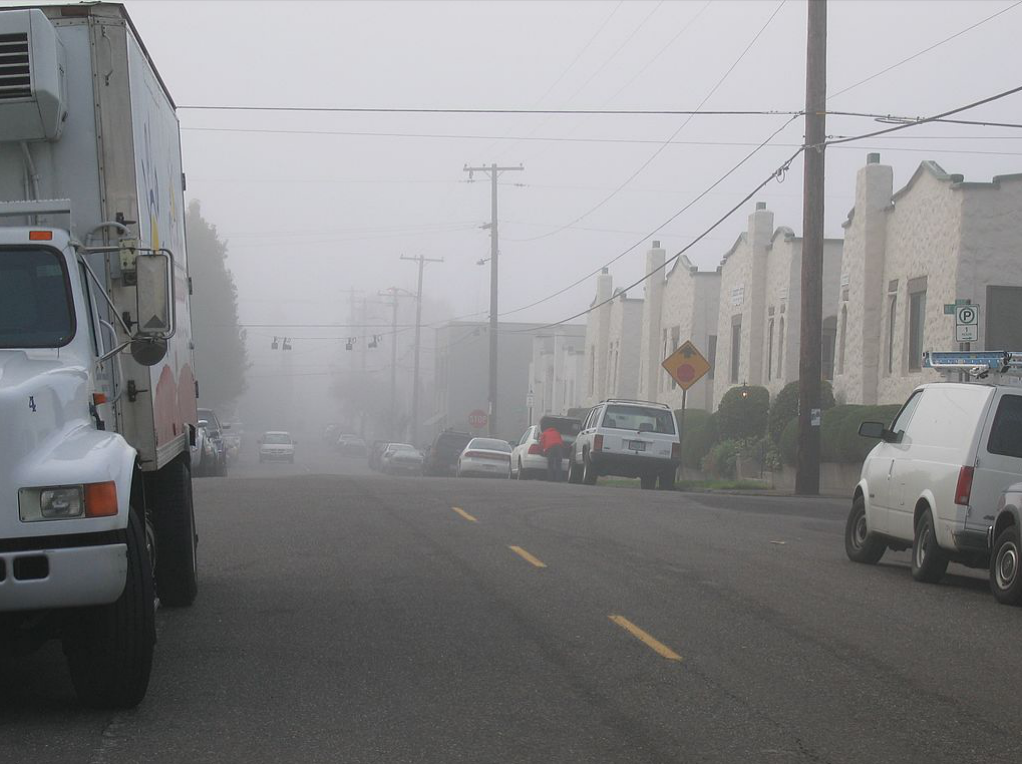}
\caption{A sample foggy image from target domain.}\label{fig:1st_exp_orig}
\end{figure}
In Table~\ref{tab:expsettings}, parameters for experiment of the proposed method are summarized. Further parameters of the experiment are analytically found out. The proposed method's training is conceptualised as the accompanying loss function's optimisation, which is carried out by training by the source domain dataset. For training and testing purposes, the hardware requirement encapsulates processor speed of 3GHz, 16GB memory working at 3200MHz, GPU supporting CUDA at 1GHz. The thermal throttling curves have been modified in accordance with the proposed method's hardware requirements.

\subsection{Results and Discussions}

\begin{figure}[t]
\centering
\includegraphics[width=0.6\linewidth]{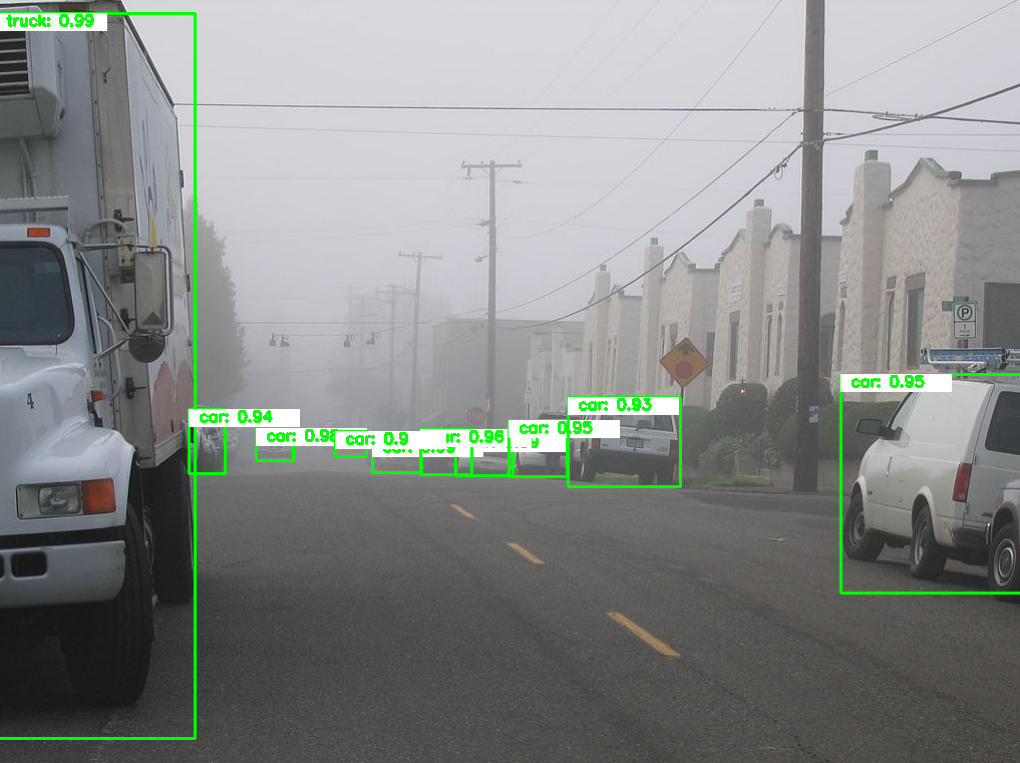}
\caption{Object detection in image of Fig.~\ref{fig:1st_exp_orig} by the proposed method.}\label{fig:1st_exp_pred}
\end{figure}
\begin{figure}[b]
\centering
\includegraphics[width=0.6\linewidth]{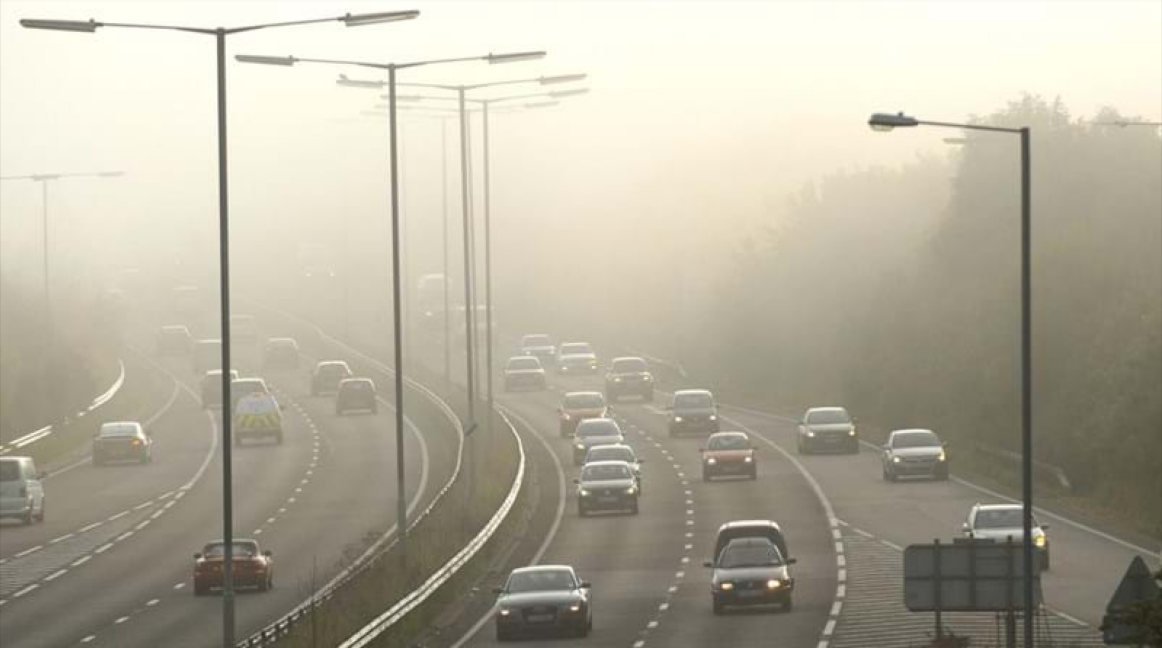}
\caption{Another sample foggy image from target domain.}\label{fig:2nd_exp_orig}
\end{figure}
\begin{figure}[t]
\centering
\includegraphics[width=0.6\linewidth]{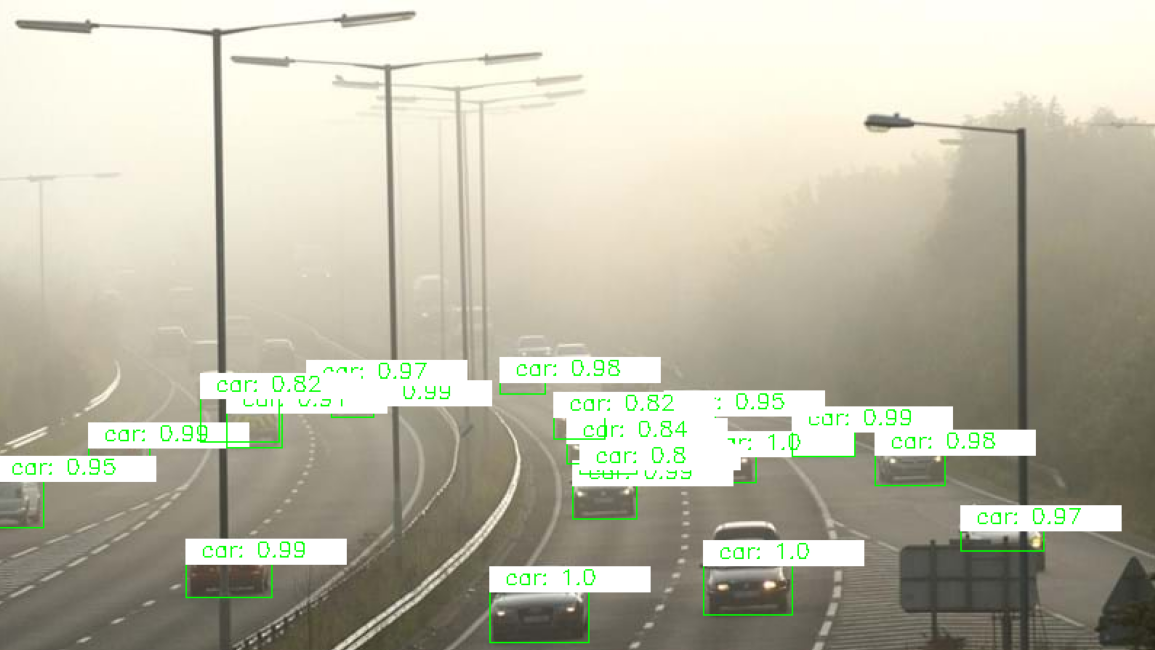}
\caption{Object detection in image of Fig.~\ref{fig:2nd_exp_orig} by the proposed method.}\label{fig:2nd_exp_pred}
\end{figure}

The outcome of the experiments depict the increase in accuracy for identifying compact bounding boxes of objects in reduced visibility. Two sample foggy images from target domain are shown in Fig.~\ref{fig:1st_exp_orig} and Fig.~\ref{fig:2nd_exp_orig}. Corresponding to the two samples of the dataset, predictions of the object classes and identifying their bounding boxes are depicted in Fig.~\ref{fig:1st_exp_pred} and Fig.~\ref{fig:2nd_exp_pred} respectively.

Table~\ref{tab:expComp} presents a comparative analysis between the nature and accuracy outcomes of the proposed method and the existing techniques on foggy image dataset. The table indicates the accuracy of roughly 85\% of the proposed method, which represents a notable increase over the existing techniques based on FerRCNNs. The proposed method outperforms the existing techniques due to significant effort on domain adaptation in the proposed method. The relevance of the source and target domains are maintained while experimentations are performed.
\begin{table}[tb]
\caption{Comparison of the proposed method to the existing works.}\label{tab:expComp}
\begin{center}
\begin{tabular}{@{}l@{~} @{$\quad$}c@{~} @{~}c@{}}
\hline
$\quad$Work & Nature & Accuracy\\%
\hline
      ~DAFerRCNN & FerRCNN\,+\,DA & 75.4\%\\
      ~FerRCNN-OpCV & FerRCNN\,+\,OpenCV & 78.4\%\\
      ~FerRCNN-DpRc & FerRCNN\,+\,DepthRecons & 82.1\%\\
      ~Proposed method & FerRCNN\,+\,DA\,+\,ResNet & 85.2\%\\
\hline
\end{tabular}
\end{center}
\end{table}

\section{Conclusion}\label{sec:concl}

This paper has presented a new foggy object detection method. The proposed method has two-staged FerRCNN architecture with loss regularization. The method is experimented on benchmark dataset of images of roads and vehicles therein under foggy weather conditions. Experiment results have confirmed notable improvements in the proposed method's accuracy over existing works. So, the proposed method paves the way for less need of emergency management~\cite{pan21}.

\bibliographystyle{unsrt}
\bibliography{citation}

\end{document}